\documentclass[runningheads]{llncs}

\usepackage{eccv}

\usepackage{eccvabbrv}

\usepackage{graphicx}
\usepackage{booktabs}

\usepackage[accsupp]{axessibility}  % Improves PDF readability for those with disabilities.
\usepackage{ulem}
\usepackage{hyperref}

\usepackage{orcidlink}
\usepackage{array}
\usepackage{booktabs}
\usepackage{tikz}

\usetikzlibrary{arrows.meta,positioning,shapes.multipart,fit}
\begin{document}

\title{GSMem: 3D Gaussian Splatting as Persistent Spatial Memory for Zero-Shot Embodied Exploration and Reasoning} 

\titlerunning{GSMem}

\author{Yiren Lu\inst{1}* \and
Yi Du\inst{2}* \and
Disheng Liu\inst{1} \and
Yunlai Zhou\inst{1} \and
Chen Wang \inst{2} \and
Yu Yin \inst{1}
}

\authorrunning{Y.~Lu et al.}

\institute{Case Western Reserve University\\
\email{\{yxl3538, dxl952, yxz3057, yxy1421\}@case.edu}\\
\and
Spatial AI \& Robotics (SAIR) Lab, University at Buffalo\\
\email{\{yid, chenw\}@sairlab.org}}
\begingroup
\renewcommand\thefootnote{}
\footnotetext{* Equal contribution.}
\endgroup

\maketitle

\begin{abstract}
Effective embodied exploration requires agents to accumulate and retain spatial knowledge over time. 
However, existing scene representations, such as discrete scene graphs or static view-based snapshots, lack \textit{post-hoc re-observability}. 
If an initial observation misses a target, the resulting memory omission is often irrecoverable. 
To bridge this gap, we propose \textbf{GSMem}, a zero-shot embodied exploration and reasoning framework built upon 3D Gaussian Splatting (3DGS). 
By explicitly parameterizing continuous geometry and dense appearance, 3DGS serves as a persistent spatial memory that endows the agent with \textit{Spatial Recollection}: the ability to render photorealistic novel views from optimal, previously unoccupied viewpoints. 
To operationalize this, GSMem employs a retrieval mechanism that simultaneously leverages parallel object-level scene graphs and semantic-level language fields. This complementary design robustly localizes target regions, enabling the agent to ``hallucinate'' optimal views for high-fidelity Vision-Language Model (VLM) reasoning. 
Furthermore, we introduce a hybrid exploration strategy that combines VLM-driven semantic scoring with a 3DGS-based coverage objective, balancing task-aware exploration with geometric coverage.
Extensive experiments on embodied question answering and lifelong navigation demonstrate the robustness and effectiveness of our framework.
  
\end{abstract}
\section{Introduction}
\begin{figure}
    \centering
    \includegraphics[width=1\linewidth]{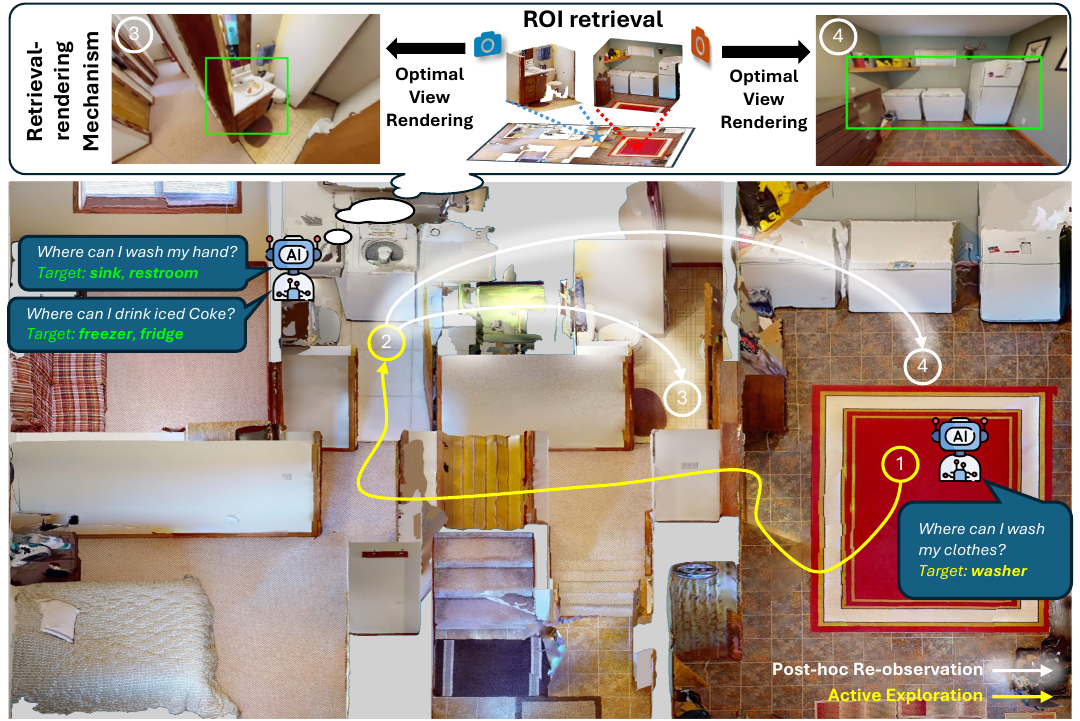}
    \caption{With GS-Mem, previously explored regions can be retrieved and re-observed directly from the 3DGS memory without physically navigating to them.}
    % \vspace{-1.5em}
    \label{fig:teaser}
\end{figure}
Embodied navigation has recently emerged as a central problem in embodied AI, driving growing interest in agents that interact with complex 3D environments.
Effective embodied exploration in such environments depends on an agent's ability to accumulate and retain spatial knowledge over time. 

% To support such capabilities, existing systems adopt different forms of scene representation. One line of work focuses on object-centric abstractions, particularly 3D scene graphs \cite{gu2024conceptgraphs, yin2024sg}, which model environments through nodes representing objects and edges capturing their relationships. Another line employs view-based representations, such as 2D top-down maps \cite{yokoyama2024vlfm, chang2023goat}, or egocentric image snapshots \cite{yang20253d}, to encode spatial observations for planning and querying.
% Despite their effectiveness, these representations do not provide comprehensive coverage of all previously explored regions. 
% Object-centric abstractions focus primarily on discrete entities and their relationships, often omitting fine-grained geometry and appearance details. Similarly, view-based representations encode observations in compressed or view-dependent forms, which cannot fully preserve the complete spatial and visual content of the environment. 
% As a result, agents may lack access to detailed information necessary for flexible reasoning and revisitation across long horizons.

To support such capabilities, existing systems adopt different forms of scene representation. 
One line of work focuses on \textit{object-centric abstractions}, such as 3D scene graphs~\cite{gu2024conceptgraphs,yin2024sg}, which model environments as a discrete set of nodes and edges representing objects and their relationships. 
Another line employs \textit{view-based representation}s, such as 2D top-down maps~\cite{yokoyama2024vlfm,chang2023goat} or egocentric image snapshots~\cite{yang20253d}, to encode spatial observations for planning and querying.

Despite their effectiveness, both paradigms lack comprehensive coverage of previously explored regions required for robust reasoning and navigation. 
\textit{Object-centric methods} rely highly on the accuracy of real-time perception modules. Since they discard raw visual data for discrete labels, a single detection failure results in an irrecoverable memory omission. Such errors propagate directly to downstream tasks, causing retrieval failures, particularly for open-vocabulary objects.
Conversely, \textit{view-based representations} remain inherently sparse and view-dependent. If a target is captured from a suboptimal angle or suffers from occlusion during the initial pass, the static memory lacks the geometric fidelity needed for VLMs to resolve ambiguity or for the agent to plan precise re-visitation.

We identify a fundamental gap underlying these limitations: the lack of \textit{post-hoc re-observability}. Unlike humans, who can mentally re-visit a past scene from a new perspective to discover missed details, current agents are ``locked'' into the specific observations they made during their initial exploration.

To bridge this gap, we propose GSMem, a zero-shot embodied exploration and reasoning framework built upon 3D Gaussian Splatting (3DGS) \cite{kerbl20233d}. 
Specifically, 3DGS explicitly parameterizes both scene geometry and dense appearance, inherently supporting real-time, high-fidelity novel view synthesis.
Leveraging this ability, we move beyond discrete abstractions and static snapshots toward a dense, continuous radiance field that serves as a persistent spatial memory. 
As shown in \cref{fig:teaser}, this allows the agent to perform spatial recollection: it can ``re-visit'' previously explored regions by rendering photorealistic views from arbitrary, optimal viewpoints, even those it never physically occupied.

% Our framework introduces a multi-level retrieval-rendering mechanism that mirrors the hierarchy of human memory. 
To ensure robust localization, our framework introduces a multi-level retrieval-rendering mechanism.
When tasked with a query, GSMem localizes the region of interest by using two parallel representations: object-level cues from a 3D scene graph and semantic-level features from a language field. These two levels strongly complement each other. 
Crucially, even if the scene graph fails to detect a specific target at the object level, the semantic-level embeddings ensure the candidate region can still be accurately localized. 
Because the underlying 3DGS memory remains intact, the agent can then ``hallucinate'' an optimal viewpoint toward this candidate region, rendering a high-fidelity observation for a VLM to perform subsequent reasoning.
Furthermore, we address the challenge of task-aware exploration by designing a hybrid strategy that combines VLM-driven semantic scoring with a 3DGS-based coverage objective. 
By quantifying the representational uncertainty of the Gaussian field via entropy, our agent can identify under-observed regions that offer the highest information gain, ensuring the constructed memory is both comprehensive and semantically rich.

In summary, our contributions are as follows:
\begin{itemize}
    \item We propose \textbf{GSMem}, a zero-shot embodied exploration and reasoning framework built upon persistent 3D Gaussian memory, which equips agents with  the capability of spatial recollection.

    \item We introduce a \textbf{multi-level retrieval-rendering mechanism} that integrates object-level scene graphs and semantic-level language fields to localize relevant regions, followed by optimal viewpoint selection for post-hoc re-observation to support VLM reasoning.

    \item We design a \textbf{hybrid exploration strategy} that combines VLM-driven semantic relevance with 3DGS-based information gain to achieve balanced and task-aware zero-shot exploration.

    \item Extensive experiments on \textbf{embodied question answering} and \textbf{lifelong navigation} demonstrate the effectiveness and generalization capability of our framework.
\end{itemize}

\section{Related Work}
\label{sec:related_work}
\bigskip
\noindent\textbf{Scene Representations for Embodied Agents.}
Effective spatial memory is fundamental for embodied agents to reason about their environment.
Early approaches largely relied on 2D top-down maps or occupancy grids, which support navigation but often discard the fine-grained visual details necessary for complex semantic reasoning.
To address this, object-centric representations, such as ConceptGraphs~\cite{gu2024conceptgraphs} and Hydra~\cite{hughes2024foundations}, were introduced to model environments as structured graphs of objects and their relationships.
While these methods provide high-level semantic abstractions, they inherently discretize the world, potentially losing geometric precision and context outside of detected objects. 
Alternatively, view-based representations, such as 3D-Mem~\cite{yang20253d} and GOAT~\cite{chang2023goat}, maintain memory as a collection of egocentric image snapshots or keyframes.
While these preserve visual fidelity, they are view-dependent and sparse, limiting the agent's ability to extrapolate novel viewpoints or reason about 3D geometry continuously.
In contrast, our GSMem utilizes 3D Gaussian Splatting~\cite{kerbl20233d} to maintain a dense, continuous, and re-renderable memory.
Unlike recent offline semantic 3DGS approaches~\cite{qin2024langsplat,zhou2024feature}, we employ an optimization-free language field that updates in real-time, enabling immediate zero-shot retrieval without the artifacts of discretization or the latency of iterative training.

\medskip
\noindent\textbf{Embodied Exploration and Navigation.}
Embodied navigation has evolved from reinforcement learning (RL) baselines to zero-shot VLM agents. 
Traditional methods often rely on trained policies or modular skill chains, such as Modular GOAT~\cite{chang2024goat} and CLIP on Wheels (CoW)~\cite{gadre2023cows}.
The GOAT-Bench benchmark~\cite{khanna2024goat} further established baselines utilizing monolithic and skill-chain policies.
However, these methods often struggle to generalize to open-vocabulary queries in unseen environments without extensive training.
The rise of VLMs has shifted the focus toward zero-shot exploration.
Methods like Explore-EQA~\cite{ren2024explore}, VLMNav~\cite{goetting2025end}, and DyNaVLM~\cite{ji2025dynavlm} utilize VLM reasoning to select frontiers or generate subgoals directly from observations. More recent approaches, such as TANGO~\cite{ziliotto2025tango} and MTU3D~\cite{zhu2025move}, leverage powerful commercial models (e.g., GPT-4o~\cite{hurst2024gpt}) to enhance reasoning and planning capabilities. Despite their advancements, these methods often rely on discrete observations or sparse graphs. Our framework distinguishes itself by combining VLM-driven semantic scores with a rigorous information-theoretic coverage objective, ensuring robust zero-shot performance that outperforms both RL-based and VLM-based baselines.
\section{Methodology}
\subsection{3DGS Mapping}
In this section, we first review the preliminaries of 3D Gaussian Splatting, and then go through the 3DGS mapping and online language field generation processes during active exploration.

\medskip
\noindent\textbf{Preliminaries of 3DGS.}
3D Gaussian Splatting (3DGS) represents a scene as a set of anisotropic 3D Gaussians $\mathcal{G}=\left\{g_i\right\}_{i=1}^N$, each defined by its mean position $\mathbf{x}_i$, covariance $\mathbf{\Sigma}_i$, opacity $\sigma_i$, and color $\mathbf{c}_i$.
To improve efficiency, we set the spherical harmonics (SH) degree to 0 and directly model color in RGB.

Given a scene represented by a set of 3D Gaussians $\mathcal{G}$, novel view rendering is achieved by projecting the Gaussians onto the image plane of a target camera and compositing them in a differentiable manner. 
The projection of a 3D Gaussian (mean $\mathbf{x}_i$, covariance $\mathbf{\Sigma}_i$) onto the image plane yields a 2D Gaussian (mean $\mathbf{x}_i'$, covariance $\mathbf{\Sigma}_i'$):
\begin{equation}
\mathbf{x}_i' = \pi\!\left(\mathbf{K}\mathbf{T}_{cw}\tilde{\mathbf{x}}_i\right),
\quad
\mathbf{\Sigma}_i' =
\mathbf{J}_i \mathbf{R}_{cw}
\mathbf{\Sigma}_i
\mathbf{R}_{cw}^{\top}
\mathbf{J}_i^{\top},
\end{equation}
where $\tilde{\mathbf{x}}_i$ denotes the homogeneous coordinate of the 3D mean $\mathbf{x}_i$,
$\mathbf{T}_{cw}=[\mathbf{R}_{cw}|\mathbf{t}_{cw}]$ is the world-to-camera transformation,
$\mathbf{K}$ is the intrinsic matrix,
$\pi(\cdot)$ represents homogeneous normalization,
and $\mathbf{J}_i$ is the Jacobian of the perspective projection evaluated at the transformed mean.

Given the projected 2D Gaussians, the final rendered color and depth at pixel $\mathbf{p}$ are obtained via standard alpha blending. 
Let $\mathcal{N}$ denote the set of Gaussians overlapping pixel $\mathbf{p}$, sorted in depth order. To facilitate our subsequent formulations, we define the blending weight $w_{i,\mathbf{p}}$ of Gaussian $g_i \in \mathcal{N}$ at pixel $\mathbf{p}$ as:
\begin{equation}
w_{i,\mathbf{p}} = \alpha_i(\mathbf{p}) \prod_{j=1}^{i-1} \left(1 - \alpha_j(\mathbf{p})\right),
\label{eq:blending-weight}
\end{equation}
where $\alpha_i(\mathbf{p})$ is the per-pixel opacity contribution derived from the learned opacity parameter $\sigma_i \in [0,1]$ and the projected 2D covariance $\mathbf{\Sigma}_i'$.
Using the blending weights $w_{i,\mathbf{p}}$, the rendered RGB color $\mathbf{C}(\mathbf{p})$ and depth $D(\mathbf{p})$ can be compactly formulated as:
\begin{equation}
\mathbf{C}(\mathbf{p}) = \sum_{i \in \mathcal{N}} w_{i,\mathbf{p}} \mathbf{c}_i, \quad
D(\mathbf{p}) = \sum_{i \in \mathcal{N}} w_{i,\mathbf{p}} z_i,
\label{eq:alpha-blending}
\end{equation}
where $z_i$ is the depth of the Gaussian center in camera coordinates.

\medskip
\noindent\textbf{Keyframe Selection \& Optimization.}
Our mapping system takes RGB-D images from sensors as input and maintains a keyframe set $\mathcal{K}$ to incrementally update the 3DGS map. 
At each time step, the agent captures three surrounding RGB-D views.
For each incoming frame, we compute the average flow magnitude between the current frame and the last keyframe using a pretrained optical flow network~\cite{teed2020raft}. 
If the average flow exceeds a predefined threshold, the frame is selected as a new keyframe and added to $\mathcal{K}$.

As exploration progresses, the number of keyframes grows. 
To ensure efficient and stable map updates, we adopt a sliding window $\mathcal{W} \subset \mathcal{K}$ with a fixed size of $10$. 
During each optimization step, we update the Gaussian map using keyframes within the current window $\mathcal{W}$, together with two additional frames randomly sampled from $\mathcal{K}$. 
Let $\mathcal{S}$ denote this sampled subset, and define the optimization set as
$\
\mathcal{T} = \mathcal{W} \cup \mathcal{S}.
$
The optimization targets are defined as
\begin{equation}
\mathcal{L}_{\text{rgb}} =
\sum_{k \in \mathcal{T}}
\left\|
\hat{\mathbf{I}}_k - \mathbf{I}_k
\right\|_1,
\quad
\mathcal{L}_{\text{depth}} =
\sum_{k \in \mathcal{T}}
\left\|
\hat{\mathbf{D}}_k - \mathbf{D}_k
\right\|_1,
\end{equation}
predicted RGB-D images $\hat{\mathbf{I}}_k$ and $\hat{\mathbf{D}}_k$ are rendered via~\cref{eq:alpha-blending} and supervised against the observed images $\mathbf{I}_k$ and $\mathbf{D}_k$ to update Gaussian parameters.

\medskip
\noindent\textbf{Online Language Field Generation.}
% Recent works have attempted to assign semantic embeddings $\mathbf{f}_i$ to each 3D Gaussian to construct a feature field. 
% However, they either optimize high-dimensional features~\cite{zhou2024feature,qin2024langsplat} or compute semantics only after full reconstruction~\cite{cheng2024occam,lu2025segment}, leading to high computational cost and limited suitability for online mapping. 
% In contrast, we propose an online, optimization-free language field generation method.
To enable semantic-level grounding, we equip each 3D Gaussian $g_i$ with a language embedding $\mathbf{f}_i$. 
Existing methods typically achieve this by either optimizing high-dimensional semantic features~\cite{zhou2024feature, qin2024langsplat} or assigning features after scene reconstruction~\cite{cheng2024occam}. 
To enable efficient online mapping, we propose an optimization-free approach that directly lifts dense 2D semantics onto 3D Gaussians.

We first extract dense 2D features $\mathbf{f}_{p, k}^{2D}$ at pixel $p$ in keyframe $k$ using a pixel-wise CLIP encoder~\cite{xie2024sed} combined with a lightweight super-resolution decoder~\cite{katragadda2025online}. 
To minimize memory overhead, these features are compressed from 768 to 32 dimensions via a pretrained autoencoder~\cite{katragadda2025online}.
Instead of relying on multi-pass rendering or additional optimization, we perform a weight-consistent reverse aggregation. 
In standard forward rendering, a 2D pixel's feature is generated by alpha-blending 3D Gaussian features. Recognizing this structural symmetry, we invert the process: we distribute the 2D pixel features back to the 3D Gaussians using the exact same blending weights. Let $w_{i,p,k}^t$ denote the blending weight of Gaussian $g_i$ at pixel $p$ in keyframe $k$ during time step $t$. 
Then the online update of the Gaussian feature $\mathbf{f}_i^t$ at time $t$ is defined as:
\begin{equation}
\begin{gathered}
\mathbf{f}_i^t =
\frac{W_i^{t-1}\mathbf{f}_i^{t-1} + \sum_{k \in \mathcal{T}_t} \sum_{p} w_{i,p,k}^t \mathbf{f}_{p,k}^{2D}}
{W_i^t}, \\
W_i^t = W_i^{t-1} + \sum_{k \in \mathcal{T}_t} \sum_{p} w_{i,p,k}^t .
\end{gathered}
\end{equation}
Since the blending weight is inherently computed during the forward RGB-D rendering process (as defined in~\cref{eq:blending-weight}), this formulation generates a dense 3D language field without introducing any optimization overhead.

In addition to the 3DGS map and language field, we maintain an object-level scene graph for structured region retrieval. 
Following ConceptGraphs~\cite{gu2024conceptgraphs}, we perform object detection, matching, and merging on the incoming RGB-D stream to obtain a consolidated set of objects with their 3D locations and semantic labels. 
For each object, we store the highest-confidence detection pose for subsequent region retrieval. 
Additionally, we maintain a frontier map and a TSDF (Truncated Signed Distance Function) map to facilitate region retrieval and exploration.

\subsection{Multi-level Retrieval-Rendering}
\begin{figure}
    \centering
    \includegraphics[width=1\linewidth]{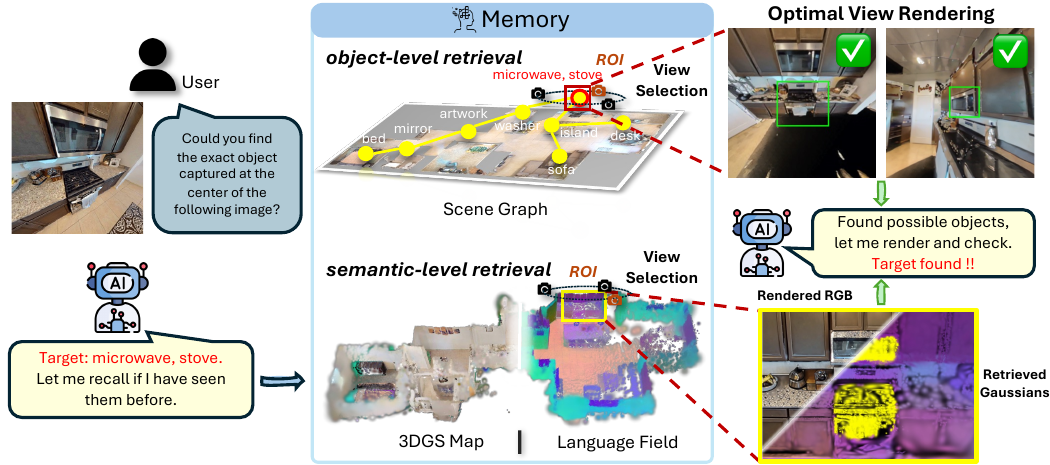}
    \caption{\textbf{Demonstration of Multi-level Retrieval-Rendering.} Given a task-related target, the agent retrieves ROIs based on object-level and semantic-level cues. Subsequent viewpoint selection and rendering enable the agent to re-observe these regions for further reasoning.}
    \label{fig:retrieval-rendering}
\end{figure}

In this section, we describe the proposed multi-level retrieval-rendering mechanism, illustrated in \cref{fig:retrieval-rendering}. 
We first extract multi-level cues from memory to retrieve task-relevant regions, and then select suitable viewpoints that allow the agent to ``revisit'' these regions for further reasoning.

\medskip
\noindent\textbf{Object-level Region Retrieval.}
At each step, we provide the VLM with the input question together with all objects maintained in the scene graph. 
The VLM then ranks the objects based on their semantic relevance to the question and selects the top-$K_\text{obj}$ candidates.
Each selected object defines an object-level region of interest (ROI) for subsequent viewpoint selection and rendering.

\medskip
\noindent\textbf{Semantic-level Region Retrieval.}
At the beginning of each episode, we provide the VLM with the input question and prompt it to identify relevant target objects or semantic entities associated with the task. 
The VLM returns a set of target descriptions, which are subsequently encoded into CLIP embeddings.
During each exploration step, we query the 3D language field with these embeddings and retrieve Gaussians whose cosine similarity exceeds a threshold $\tau_{\text{clip}}$.

Since the retrieved Gaussians may span multiple regions, we further cluster them into spatially coherent groups.
We use a KD-Tree~\cite{maneewongvatana1999analysis} to identify neighbors within distance $\tau_d$ and construct an adjacency graph. 
Connected components are treated as candidate clusters, from which we discard small clusters and retain the top-$K_{\text{cluster}}$ ranked by mean cosine similarity as semantic-level ROIs.

% \subsubsection{Viewpoint Selection for Visual Recall.}
\medskip
\noindent\textbf{Optimal Rendering Viewpoint for Visual Recall.}
After obtaining the object-level and semantic-level ROIs, the agent must determine the optimal viewpoint to render each region for VLM reasoning. 
Our overall viewpoint selection strategy follows a sample-then-score paradigm: 
we first sample a dense set of candidate camera poses around the target, and then evaluate them through a two-phase scoring process to find the optimal view.

Specifically, given the 3D bounding box of an ROI, we sample candidate viewpoints along a horizontal circular trajectory centered at the ROI.  
By uniformly sampling 36 azimuth angles at $10^\circ$ intervals and three elevation angles ($-10^\circ, 0^\circ, 15^\circ$), we effectively generate $108$ candidate poses per ROI. These candidates are then filtered and ranked as follows:

\noindent \textbf{Phase 1}:
As a preliminary step, we perform coarse feasibility filtering by discarding any camera poses that lie inside obstacle regions of the TSDF map. 
We then evaluate the remaining candidates based on a ray visibility score $S_{\text{vis}}$ and a projected area score $S_A$.

For visibility, we sample 14 representative points (8 vertices and 6 face centers) on the ROI's 3D bounding box and perform ray marching on the TSDF map to check their visibility from the candidate viewpoint. $S_{\text{vis}}$ is defined as the ratio of visible points to the total 14 points. 
For coverage, we project the 8 bounding box vertices onto the candidate view to compute 2D area $A$. 
To encourage an appropriate viewing scale, avoiding views that are too close (limited context) or too far (loss of detail), we define $S_A$ using a Gaussian penalty:

\begin{equation}
S_{\text{vis}} = \frac{N_{\text{visible}}}{14},
\quad
S_A =
\exp\left(
-\frac{(A - A^*)^2}{2{\sigma_a}^2}
\right),
\label{s_vis and s_a}
\end{equation}
where $A^*$ is the target projected area and $\sigma_a$ controls tolerance to scale deviation.
Candidate poses are ranked by the combined score $S_{\text{vis}} + S_A$, and the top-10 poses are advanced to the final phase.

\noindent \textbf{Phase 2}:  
For the top-10 candidate viewpoints, we further assess the actual rendering quality using the 3DGS opacity map. 
We render an opacity map for each pose and compute the average opacity within the projected 2D ROI. 
Due to the compositional nature of 3DGS, a higher accumulated opacity indicates a stronger surface presence and better visual observability. 
The opacity score $S_{\text{opa}}$ is defined as:

\begin{equation}
S_{\text{opa}} = \frac{1}{|\Omega|} \sum_{p \in \Omega} \alpha_p,
\end{equation}
where $\Omega$ is the set of pixels within the projected ROI , and $\alpha_p$ is the accumulated opacity value at pixel $p$.
Finally, we select the candidate with the highest combined score as the optimal rendering viewpoint for visual recall: 
\begin{equation}
S_{\text{final}} = S_{\text{vis}} + S_A + S_{\text{opa}}.
\end{equation}

\subsection{ROI rendering \& Vision-Language Reasoning}
Given the optimal viewpoint selected in the previous stage, we render the ROI from this viewpoint for visual analysis. 
In addition, we also render the scene from the highest-confidence detection pose of the corresponding object to provide complementary visual evidence.
To further improve the rendering quality under novel viewpoints, we incorporate a single-step diffusion model~\cite{wu2025difix3d+} to enhance the visual fidelity of the rendered images before feeding them into the VLM.

The rendered views are then fed into the VLM to determine whether the question can be answered with the current observations. 
The detailed prompting strategy is provided in the supplementary material.
If the VLM determines that the available visual evidence is insufficient, the agent proceeds to further explore the environment following our Hybrid Exploration Strategy.

\subsection{Hybrid Exploration Strategy}
\begin{figure}
    \centering
    \includegraphics[width=1\linewidth]{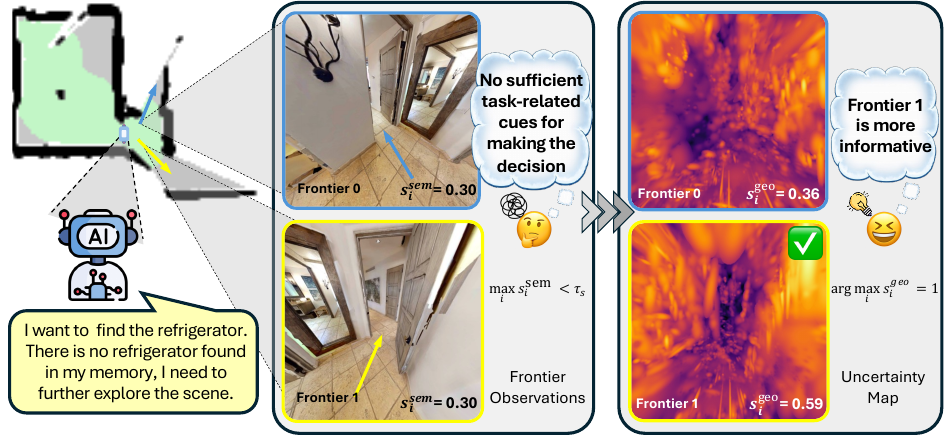}
    \caption{\textbf{Demonstration of our Hybrid Exploration Strategy.} When frontier observations do not contain sufficient task-related cues for the VLM to make a decision, we incorporate an information gain-based score to select the most informative frontier for further exploration.}
    \label{fig:hybrid-exploration}
\end{figure}
Our exploration module follows the classical frontier-based paradigm~\cite{yamauchi1997frontier}, 
where frontiers are defined as the boundaries between known free space and unexplored regions in the occupancy map. 
At each exploration step, we extract a set of candidate frontiers from the 2D occupancy map and evaluate them using a hybrid criterion that balances semantic relevance and geometric coverage.

\medskip
\noindent\textbf{Semantic Relevance.}
For each candidate frontier, we feed the corresponding observation to a vision-language model (VLM) conditioned on the task query. 
The VLM outputs a normalized relevance score $s_i^{\text{sem}} \in [0,1]$, reflecting how likely the frontier is to reveal information useful for answering the question.

\medskip
\noindent\textbf{Geometric Coverage.}
To promote exploration toward geometrically informative regions, we score each frontier $\xi_i$ (with an associated camera pose $\mathbf{T}_i$) based on the expected information gain of acquiring new observations $\mathcal{Y}_i$. 
Following information-theoretic formulations~\cite{lindley1956measure}, this corresponds to the reduction in the differential entropy $\mathcal{H}$ of the 3DGS parameters $\theta$ given the active window $\mathcal{W}_t$:
\begin{equation}
    s_i^{\text{geo}} = \mathcal{H}(\theta \mid \mathcal{W}_t) - \mathcal{H}(\theta \mid \mathcal{W}_t, \mathcal{Y}_i).
\end{equation}

Under the Laplace approximation, maximizing this information gain is equivalent to the D-optimality criterion, which computes the log-determinant of the Fisher Information Matrix (FIM)~\cite{houlsby2011bayesian,kirsch2019batchbald,jiang2023fisherrf}:
\begin{equation}
    \mathcal{I}(\theta; \mathcal{Y}_i \mid \mathcal{W}_t) \approx \frac{1}{2}\log\det\!\left(\mathbf{I}_t+\mathbf{I}_i\right) - \frac{1}{2}\log\det\!\left(\mathbf{I}_t\right),
\end{equation}
where $\mathbf{I}_t$ and $\mathbf{I}_i$ respectively quantify the prior geometric certainty from $\mathcal{W}_t$ and the new information introduced by $\xi_i$, both approximated via the inner product of the rendering Jacobians (\ie, $\mathbf{J}^\top \mathbf{J}$). 
Because evaluating the determinant of the high-dimensional global FIM $\mathbf{I}_t$ is computationally intractable for 3DGS, we adopt an efficient T-optimality surrogate. We approximate the geometric score using the trace of the incremental FIM $\mathbf{I}_i$:
\begin{equation}
    s_i^{\text{geo}} \approx \mathrm{Tr}(\mathbf{I}_i).
\end{equation}
This trace-based metric acts as an efficient proxy for geometric uncertainty, which can be computed directly from the gradients of the 3DGS rendering function without ground-truth supervision.

\medskip
\noindent\textbf{Exploration Strategy.}
Based on the two scores defined above, we design the following hybrid exploration strategy, as shown in \cref{fig:hybrid-exploration}.
At each step, if $\max_i s_i^{\text{sem}} > \tau_s$, we select the frontier with the highest semantic score. 
Otherwise, when no frontier provides sufficient task-relevant information for the VLM, we select the frontier that maximizes the geometric exploration score:
\begin{equation}
i^* =
\begin{cases}
\displaystyle \arg\max_i \; s_i^{\text{sem}},
& \text{if } \max_i s_i^{\text{sem}} > \tau_s, \\[6pt]
\displaystyle \arg\max_i \; s_i^{\text{geo}},
& \text{otherwise}.
\end{cases}
\end{equation}

\section{Experiments}
To systematically evaluate GSMem, we conduct experiments from two complementary perspectives. 
First, we evaluate Active Embodied Question Answering (A-EQA), where the agent starts in an unseen environment and must actively explore to gather evidence for answering task queries, testing the general exploration and reasoning ability of GSMem. 
Second, we consider a Multimodal Lifelong Navigation setting to assess memory reliability over long-horizon interactions. 
In this setting, the agent continuously accumulates observations across episodes, and targets may appear in previously explored regions, requiring the system to retrieve and utilize stored memory effectively.

\subsection{Active Embodied Question Answering}
\label{sec:aeqa}
\noindent\textbf{Benchmark.}
We evaluate GSMem on the Active Embodied Question Answering (A-EQA) benchmark introduced in OpenEQA~\cite{majumdar2024openeqa}. 
It contains question-answer pairs from 63 scenes in Habitat-Matterport3D (HM3D)~\cite{ramakrishnan2021habitat}, covering tasks such as object recognition, functional reasoning, and spatial understanding. 
At the start of each episode, the agent is placed in an unexplored environment and needs to answer the question through active exploration.
Following~\cite{yang20253d}, we use the 184-question evaluation set released by OpenEQA.

\smallskip
\noindent\textbf{Metrics.}
Following prior work~\cite{majumdar2024openeqa,yang20253d}, we adopt LLM-Match and LLM-Match SPL for evaluation. 
LLM-Match is scored by GPT-4~\cite{achiam2023gpt} on a 1–5 scale and rescaled to 0–100, while LLM-Match SPL further accounts for exploration efficiency by weighting the score with the exploration path length.

\smallskip
\noindent\textbf{Baselines.}
\label{baseline}
Following prior work~\cite{majumdar2024openeqa,zhu2025move,yang20253d}, we compare GS-Mem with four categories of baselines: 
(1) \textbf{Blind LLMs}: LLMs~\cite{hurst2024gpt,grattafiori2024llama} answer the question using only textual input. 
(2) \textbf{LLM + Captions}: Frame captions generated by LLaVA-1.5~\cite{liu2024improved} are provided to the LLM. 
(3) \textbf{VLMs}: Observations are directly fed into VLMs~\cite{hurst2024gpt,yang2025qwen3}. 
(4) \textbf{VLM Exploration}: Explore-EQA~\cite{ren2024explore}, ConceptGraphs~\cite{gu2024conceptgraphs}, and 3D-Mem~\cite{yang20253d}. 
For fairness, all methods in VLM exploration use GPT-4o~\cite{hurst2024gpt} via the OpenAI API as the VLM backbone.

\begin{table}[ht]
    \centering
    \caption{\textbf{Experiments on A-EQA.} ``CG'' denotes ConceptGraphs.
    All the VLM exploration methods use GPT-4o as VLM backbone.
    }
    \resizebox{0.7\linewidth}{!}{
    \begin{tabular}{llcc}
        \toprule
        \textbf{} & \textbf{Method} & \textbf{LLM-Match $\uparrow$} & \textbf{LLM-Match SPL $\uparrow$} \\
        \midrule
        % \multirow{2}{*}{Blind LLMs}
        & \textbf{\textit{Blind LLMs}} \\
        & GPT-4o & 35.9 & - \\
        & LLaMA-3.1-8B-Instruct & 32.2 & - \\
        \midrule
        %\multirow{2}{*}{MultiFrame VLM}
        & \textbf{\textit{LLM with Frame Captions}} \\
        & GPT-4o w/ LLaVA-1.5-13B & 33.11 & - \\
        & LLaMA-3.1 w/ LLaVA-1.5-13B & 30.84 & - \\
        \midrule
        & \textbf{\textit{VLMs}} \\
        & GPT-4o & 47.6 & - \\
        & Qwen3-VL-8B-Instruct & 44.0 & - \\
        % & Qwen3-VL-30B-A3B & 50.1 & N/A \\
        \midrule
        % \multirow{3}{*}{GPT-4o Exploration} 
        & \textbf{\textit{VLM Exploration}} \\
        %& Snapshots (observe) & 52.45 & 36.10 \\
        & Explore-EQA \cite{ren2024explore} & 46.9 & 23.4 \\
        & CG \cite{gu2024conceptgraphs} w/ Frontier Snapshots & 47.2 & 33.3 \\
        & 3D-Mem \cite{yang20253d} & 52.6 & 42.0 \\
        & GSMem (Ours) & \textbf{55.4} & \textbf{43.8} \\
        \bottomrule
    \end{tabular}
    }
    \label{tab:a-eqa}
    
\end{table}
\smallskip
\noindent\textbf{Results.}
We report the quantitative results of active EQA in \cref{tab:a-eqa}. 
GSMem achieves state-of-the-art performance on this benchmark.
Compared to high-level graph-based methods, GSMem maintains dense geometric and visual representations, providing richer visual evidence for downstream reasoning. 
Compared to the keyframe-based 3D-Mem, GSMem is able to select optimal viewpoints for retrieved regions, leading to improved visual coverage and higher-quality renderings. 
This enhanced visual evidence provides more informative inputs for the VLM, leading to stronger reasoning performance.
At the same time, our hybrid exploration strategy also improves efficiency during active exploration.

\subsection{Multi-modal Lifelong Navigation}
\noindent\textbf{Benchmark.}
GOAT-Bench~\cite{khanna2024goat} is a multimodal lifelong navigation benchmark that evaluates an agent’s ability to sequentially locate multiple targets within an unseen scene. 
Targets are described using either a semantic category name, a detailed language description, or an image of the target object.
We conduct experiments on the \textbf{valid unseen} split of GOAT-Bench, which contains 36 scenes. 
Each scene includes 10 episodes, and each episode is composed of multiple subtasks, resulting in more than 2,600 subtasks in total.

\smallskip
\noindent\textbf{Metrics.}
GOAT-Bench adopts Success Rate (SR) and Success weighted by Path Length (SPL) as evaluation metrics.
A task is considered successful if the agent’s final position is within 1 meter of the target location. 
SPL further incorporates path efficiency by weighting the success indicator according to the ratio between the shortest-path distance and the actual trajectory length.

\smallskip
\noindent\textbf{Baselines.}
We include four reinforcement learning–based baselines from the GOAT-Bench paper. 
In addition, we compare against several VLM-based exploration methods.
VLMNav~\cite{goetting2025end} and DyNaVLM~\cite{ji2025dynavlm} employ open-source VLM backbones, while recent methods such as TANGO~\cite{ziliotto2025tango}, MTU3D~\cite{zhu2025move}, and 3D-Mem~\cite{yang20253d} utilize commercial models (\ie, GPT-4o) as their VLM backbone.
To ensure a fair comparison, we also adopt GPT-4o as the underlying VLM.

\smallskip
\noindent\textbf{Results.}
The quantitative results on GOAT-Bench are presented in \cref{tab:goat-bench}.
Under the lifelong navigation setting, GSMem achieves a larger performance gain than in the A-EQA benchmark, suggesting that our persistent memory representation is particularly beneficial for long-horizon scenarios.
Moreover, the hybrid exploration strategy enables more efficient exploration, as reflected in the improved SPL metric.
\begin{table}[htbp]
    \centering
    \caption{\textbf{Experiments on  GOAT-Bench ``Val Unseen" split.}}
    \resizebox{0.6\linewidth}{!}{
    \begin{tabular}{llcc}
        \toprule
        \textbf{} & \textbf{Method} & \textbf{Success Rate $\uparrow$} & \textbf{SPL $\uparrow$} \\
        \midrule
        %\multirow{4}{*}{Other Baselines} 
        & \textbf{\textit{GOAT-Bench Baselines}} \\
        & Modular GOAT \cite{chang2024goat} & 24.9 & 17.2 \\
        & Modular CLIP on Wheels \cite{gadre2023cows} & 16.1 & 10.4 \\
        & SenseAct-NN Monolithic \cite{khanna2024goat} & 12.3 & 6.8 \\
        & SenseAct-NN Skill Chain \cite{khanna2024goat} & 29.5 & 11.3 \\
        
        \midrule
        & \textbf{\textit{VLM Exploration}} \\
        & VLMnav \cite{goetting2025end} & 20.1 & 9.6 \\
        & DyNaVLM \cite{ji2025dynavlm} & 25.5 & 10.2 \\
        & TANGO \cite{ziliotto2025tango} & 32.1 & 16.5 \\
        & MTU3D \cite{zhu2025move} & 47.2 & 27.7 \\
        & 3D-Mem \cite{yang20253d} & 62.9 & 44.7 \\
        & GSMem (Ours) & \textbf{67.2} & \textbf{46.9} \\
        \bottomrule
    \end{tabular}
    }
    \label{tab:goat-bench}
\end{table}

\begin{figure}
    \centering
    \includegraphics[width=1\linewidth]{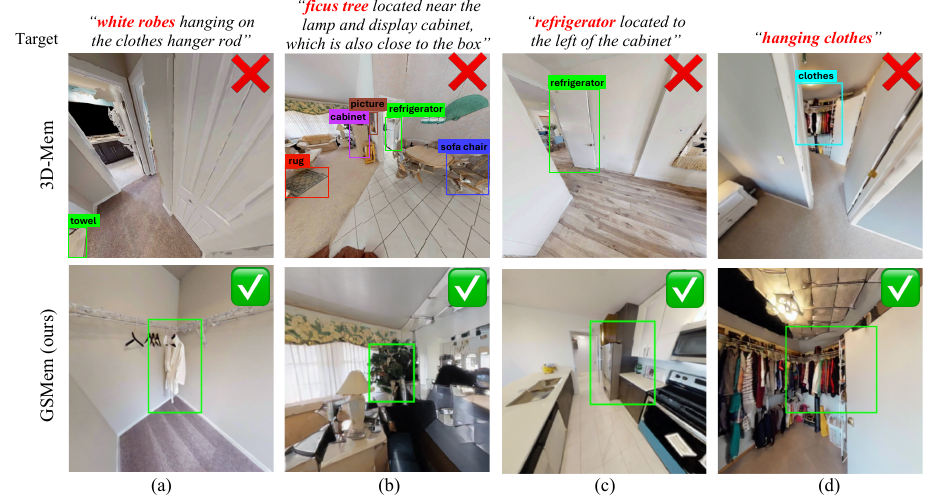}
    \caption{\textbf{Case analysis.} 
    We analyze several cases where scene-graph and view-based representations fail, and demonstrate the advantages of 3DGS-based memory. 
    The images shown correspond to the views selected by the VLM for answering the questions. 
    The examples (a-c) illustrate failures of the scene-graph detector, while the last example (d) highlights how optimal viewpoint rendering benefits the VLM's reasoning.}
    \label{fig:case_study}
\end{figure}
\subsection{Case Analysis}
\label{subsec:case_study}
% \vspace{-1em}
We present several case studies in \cref{fig:case_study} to evaluate the robustness of GSMem against the inherent limitations of discrete and view-dependent memories.

\smallskip
\noindent\textbf{Missing and Incorrect Detections.} 
A primary failure mode of current models is their heavy reliance on the performance of object detectors. 
In the first example (\cref{fig:case_study}a), the target (``white robe'')  is missed during the mapping phase. For traditional baselines, this results in an irrecoverable omission, as the object was never explicitly identified and labeled.
In the second example (\cref{fig:case_study}b), while a snapshot-based method may identify a relevant view via contextual cues, the failure of target (``ficus tree'') detection still prevents precise coordinate grounding and navigation.
Furthermore, \cref{fig:case_study}c highlights a semantic failure where the baseline misidentifies a ``white door'' as a ``refrigerator'', which subsequently misleads the VLM’s decision.
In all three cases, GSMem bypasses these perception gaps by querying its continuous language field. 
By leveraging semantic similarity for retrieval, GSMem enables correct localization and navigation even when explicit detections are missing or erroneous.

\smallskip
\noindent\textbf{View-Dependency and Resolution Limits.} The fourth example (\cref{fig:case_study}d) demonstrates the necessity of optimal viewpoint selection and rendering. Although the baseline captures the correct region, its static egocentric snapshots suffer from limited resolution and suboptimal perspective, preventing the VLM from recognizing the hanging clothes. By leveraging the re-observability nature of 3DGS, GSMem re-visits the region with an optimal viewpoint. This provides the VLMs with better visual evidence for reasoning and decision-making.

% In the first example, the agent needs to find the white robe. 
% 3D-Mem fails because the object detector cannot recognize the ``white robe.'' 
% Although GSMem uses the same detector and thus lacks the object in the scene graph, our language field retrieves the robe via semantic similarity, enabling correct localization.

% In the second example, 3D-Mem fails to detect a ficus tree. While the VLM selects a relevant snapshot based on contextual cues, the missing detection prevents accurate navigation. 
% In contrast, GSMem retrieves the ficus tree through the language field and successfully reaches the target.

% In the third example, 3D-Mem mistakenly identifies a white door as a refrigerator, misleading the VLM. 
% GS-Mem instead retrieves the correct refrigerator via the language field and provides a more representative rendering.

% Finally, although 3D-Mem captures the correct snapshot, its limited resolution prevents the VLM from recognizing the hanging clothes. 
% GS-Mem selects an optimal viewpoint with clearer visual evidence, enabling correct identification.

\subsection{Ablation Study}

We conduct ablation experiments on 10\% of the GOAT-Bench ``valid unseen'' split. 
We first evaluate the impact of key components in GSMem, including the CLIP language field, viewpoint selection, single-step diffusion enhancement, and the hybrid exploration strategy in \cref{tab:combined_ablation_module}.
Removing the language field leads to a significant drop in success rate, confirming its role as a crucial complement to object detection for open-vocabulary retrieval. 
Without the hybrid exploration strategy, relying solely on VLM-driven exploration substantially reduces SPL, highlighting the importance of coverage-aware exploration. 
Viewpoint selection improves visual grounding by providing better-aligned renderings, while the single-step diffusion model further enhances rendering quality, both contributing to improved success rates.

We further study the sensitivity to two key hyperparameters: the semantic threshold $\tau_s$ used in hybrid exploration and the top-$K_{obj}$ selected for rendering. 
The results are summarized in \cref{tab:combined_ablation_param}.
For the hybrid exploration threshold, A large $\tau_s$ biases exploration toward geometric coverage and delays target discovery.
Conversely, a small $\tau_s$ overemphasizes semantic guidance and may cause premature commitment to task-relevant regions, resulting in reduced overall success rate.
For $K_{obj}$, too few regions provide insufficient context for the VLM, whereas too many introduce redundant visual evidence. 
We therefore set $K_{obj}=10$.
\begin{table}[htbp]
\centering
\begin{minipage}{0.48\linewidth}
\centering
\small
\caption{Ablation study on different modules and designs.}
\begin{tabular}{lcc}
\toprule
\textbf{Method} & \textbf{Success Rate} $\uparrow$ & \textbf{SPL} $\uparrow$ \\
\midrule
w/o CLIP & 66.6 & 45.3 \\
w/o view selection & 68.4 & 46.3 \\
w/o diffusion & 69.8 & 48.2 \\
w/o hybrid explore & 69.9 & 47.8 \\
GSMem (Ours) & \textbf{71.1} & \textbf{51.9} \\
\bottomrule
\label{tab:combined_ablation_module}
\end{tabular}
\end{minipage}
\hfill
\begin{minipage}{0.48\linewidth}
\centering
\small
\caption{Hyperparameter ablation study}
\scalebox{0.85}{
\begin{tabular}{lcc}
% \begin{tabular}{l>{\centering\arraybackslash}p{2.5cm}c}
\toprule
\textbf{Param} & \textbf{Success Rate} $\uparrow$ & \textbf{SPL} $\uparrow$ \\
\midrule
$\tau_s=0.2$ & 69.3 & 50.8 \\
$\tau_s=0.4$ & 71.1 & 51.9 \\
$\tau_s=0.6$ & 70.6 & 49.6 \\
\midrule
$K_{obj}=4$ & 68.8 & 48.6 \\
$K_{obj}=6$ & 67.6 & 48.9 \\
$K_{obj}=10$ & 71.1 & 51.9 \\
$K_{obj}=12$ & 70.3 & 49.7 \\
\bottomrule
\label{tab:combined_ablation_param}
\end{tabular}}
\end{minipage}
\vspace{-1em}
\end{table}

\subsection{Real-world Practicality}
\begin{figure}
    \centering
    \includegraphics[width=1\linewidth]{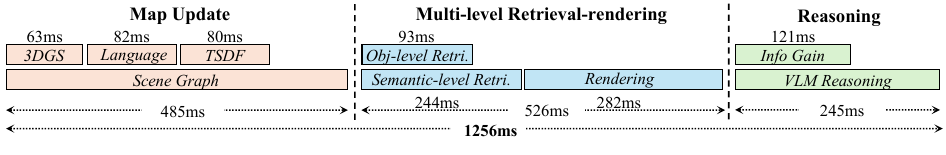}
    \caption{\textbf{Runtime analysis for real-world multi-process setup.}}
    \label{fig:runtime_analysis}
\end{figure}

We analyze the runtime of each component in our framework, and the results are shown in \cref{fig:runtime_analysis}. 
We account for modules that can operate concurrently and estimate the effective runtime of the system in a practical setting. 
For this estimation, we replace the OpenAI API with an offline-deployed Qwen3-VL-8B model and remove the single-step diffusion process during rendering. 
The navigation framework runs on a single RTX 4090, while the Qwen3 model is deployed on an H100 GPU. 
Under this multi-process setup, the average processing time per step is around 1.2 seconds.

\section{Conclusion}
In this work, we present \textbf{GSMem}, a zero-shot embodied exploration and reasoning framework built upon 3D Gaussian Splatting that equips embodied agents with \textit{post-hoc re-observability}. 
This capability allows the agent to revisit previously explored regions from arbitrary viewpoints without physically navigating to them, enabling more efficient memory retrieval and precise recall in lifelong navigation scenarios. 
As a result, GSMem achieves higher success rates and more efficient navigation compared to existing methods.


% \clearpage
\bibliographystyle{splncs04}
\bibliography{main}
\end{document}